\documentclass{bmvc2k}

\usepackage[table]{xcolor}
\usepackage{multirow}
\usepackage{subfigure}
\usepackage{xcolor}
\usepackage{fixltx2e}
\usepackage[first=0,last=9]{lcg}
\usepackage{gensymb}
\usepackage[percent]{overpic}

\newcommand{\squeezeup}{\vspace{-2mm}}
\newcommand{\squeezedown}{\vspace{+2mm}}

\newcommand{\tsb}{\textsubscript}
\newcommand{\x}{$\times$}
\newcommand{\XNET}{MRCNet}

\xdefinecolor{sport}{RGB}{255,0,0}
\xdefinecolor{city_center}{RGB}{0,255,0}
\xdefinecolor{fairs}{RGB}{255,127,80}
\xdefinecolor{festivals}{RGB}{255,192,203}
\xdefinecolor{C1}{RGB}{238,130,238}
\xdefinecolor{C2}{RGB}{255,192,203}
\xdefinecolor{C3}{RGB}{216,191,216}
\xdefinecolor{C4}{RGB}{175,238,238}
\xdefinecolor{C5}{RGB}{65,105,225}
\xdefinecolor{C6}{RGB}{30,144,255}
\xdefinecolor{C7}{RGB}{135,206,250}
\xdefinecolor{C8}{RGB}{210,105,30}
\xdefinecolor{C9}{RGB}{220,20,60}
\xdefinecolor{C10}{RGB}{250,235,215}
\xdefinecolor{C11}{RGB}{154,205,50}
\xdefinecolor{C12}{RGB}{255,255,0}
\xdefinecolor{C13}{RGB}{50,205,50}
\xdefinecolor{C14}{RGB}{144,238,144}
\xdefinecolor{C15}{RGB}{46,139,87}
\xdefinecolor{C16}{RGB}{85,107,47}
\xdefinecolor{rowColor}{RGB}{124,252,0}

\newcommand{\datasetDetailsTable}{
\begin{table}[p]
\centering
\resizebox{.95\textwidth}{!}{\begin{tabular}{|c|c|c|c|c|c|c|} 
 \hline
   Images   & Scenes                             		& Campaigns 	& GSD (cm/pixels)	& Size (pixels)		& Person Count	&Train/Test 		\\  \hline\hline
  I\tsb{1}& \cellcolor{sport}               		& \cellcolor{C1}ALZ 2015   & 11.0		&2350\x1545	&3,911  &Test	\\ \cline{1-2}\cline{4-7}
  I\tsb{2}& \cellcolor{sport} Sport      		& \cellcolor{C2}ALZ 2012 	& 14.0	 	&3744\x5616	&5,627	&Train	\\ \cline{1-2}\cline{4-7}
  I\tsb{3}& \cellcolor{sport}              		& \cellcolor{C3}BOR 2013   & 7.5 		&5619\x3744	&11,834	&Train	\\ \hline
  I\tsb{4}& \cellcolor{city_center}      		& \cellcolor{C4}MUC 2011	& 12.0 		&3744\x5616	&2,776	&Test	\\ \cline{1-2}\cline{4-7}
  I\tsb{5}& \cellcolor{city_center} City		& \cellcolor{C5}MUC 2016	& 9.0  		&3456\x5184	&3,188	&Train			\\ \cline{1-2}\cline{4-7}
  I\tsb{6}& \cellcolor{city_center} Centers	& \cellcolor{C6}MUC 2017	& 10.0 	 	&3456\x5184	&1,522	&Train			\\ \cline{1-2}\cline{4-7}
  I\tsb{7}& \cellcolor{city_center}			& \cellcolor{C6}	& 13.0  		&3744\x5616	&2,440	&Train  		\\ \cline{1-2}\cline{4-7}
  I\tsb{8}& \cellcolor{city_center}			&  \cellcolor{C7}BRA 2017	& 14.0		&3456\x5184	&478	&Train			\\ \cline{1-2}\cline{4-7}
  I\tsb{9}& \cellcolor{city_center}			& \cellcolor{C7}	& 7.0 			&3456\x5184	&508	&Test			\\ \cline{1-2}\cline{4-7}
I\tsb{10}& \cellcolor{city_center}			& \cellcolor{C7}	& 7.0	 		&3456\x5184	&285	&Test	\\ \cline{1-2}\cline{4-7}
I\tsb{11}& \cellcolor{city_center}			& \cellcolor{C7}	& 7.0			&3456\x5184	&519	&Test			\\ \cline{1-2}\cline{4-7}
I\tsb{12}& \cellcolor{city_center}			& \cellcolor{C7}	& 7.0			&3456\x5184	&448	&Test	\\ \hline
I\tsb{13}& \cellcolor{fairs}				& \cellcolor{C8}BAU 2016	& 12.0		&3456\x5184	&8,820	&Test			\\  \cline{1-2}\cline{4-7}
I\tsb{14}& \cellcolor{fairs} Fairs			& \cellcolor{C8}	& 12.0		&3456\x5184	&19,286	&Test			\\  \cline{1-2}\cline{4-7}
I\tsb{15}& \cellcolor{fairs}				& \cellcolor{C8}	& 12.0		&3456\x5184	&12,007	&Train 			\\  \cline{1-2}\cline{4-7}
I\tsb{16}& \cellcolor{fairs}				& \cellcolor{C9}WIE 2015	& 15.0 		&3744\x5616	&3,845	&Test		\\  \cline{1-2}\cline{4-7}
I\tsb{17}& \cellcolor{fairs}				& \cellcolor{C10}WIE 2016	& 13.0		&3744\x5616	&2,696	&Train		\\  \cline{1-2}\cline{4-7}
I\tsb{18}& \cellcolor{fairs}				& \cellcolor{C10}	& 13.0		&3744\x5616	&3,759	&Train		\\ \hline
I\tsb{19}& \cellcolor{festivals}				& \cellcolor{C11}RAR 2013	& 9.0 			&3744\x5616	&24,368	&Train 		\\ \cline{1-2}\cline{4-7}
I\tsb{20}& \cellcolor{festivals}Festivals		& \cellcolor{C11}	& 9.0 			&3744\x5616	&20,231	&Train		\\ \cline{1-2}\cline{4-7}
I\tsb{21}& \cellcolor{festivals}				& \cellcolor{C12}RAR 2014	& 9.0 			&3744\x5616	&6,350	&Test 		\\ \cline{1-2}\cline{4-7}
I\tsb{22}& \cellcolor{festivals}				& \cellcolor{C13}STH	2016& 4.5	 	&3744\x5616	&5,535	&Train 		\\ \cline{1-2}\cline{4-7}
I\tsb{23}& \cellcolor{festivals}				& \cellcolor{C13}	& 4.5 		&3744\x5616	&7,848	&Train 		\\ \cline{1-2}\cline{4-7}
I\tsb{24}& \cellcolor{festivals}				& \cellcolor{C13}	& 4.5		&3744\x5616	&10,083	&Train 		\\ \cline{1-2}\cline{4-7}
I\tsb{25}& \cellcolor{festivals}				& \cellcolor{C13}	& 6.5 		&3744\x5616	&5,549	&Test		\\ \cline{1-2}\cline{4-7}
I\tsb{26}& \cellcolor{festivals}				& \cellcolor{C13}	& 6.5 		&3744\x5616	&14,281	&Test 		\\ \cline{1-2}\cline{4-7}
I\tsb{27}& \cellcolor{festivals}				& \cellcolor{C14}OAC 2013	& 9.0 			&3744\x5616	&15,636	&Test		\\ \cline{1-2}\cline{4-7}
I\tsb{28}& \cellcolor{festivals}				& \cellcolor{C15}OAC 2016	& 5.5	 	&3456\x5184	&3,223	&Train 		\\ \cline{1-2}\cline{4-7}
I\tsb{29}& \cellcolor{festivals}				& \cellcolor{C15}	& 5.5 		&3456\x5184	&5,347	&Train 		\\ \cline{1-2}\cline{4-7}
I\tsb{30}& \cellcolor{festivals}				& \cellcolor{C15}	& 5.5 		&3456\x5184	&3,972	&Train 		\\ \cline{1-2}\cline{4-7}
I\tsb{31}& \cellcolor{festivals}				& \cellcolor{C15}	& 7.5 		&3456\x5184	&5,926	&Test		\\ \cline{1-2}\cline{4-7}
I\tsb{32}& \cellcolor{festivals}				& \cellcolor{C16}WIT 2017	& 11.0 		&3456\x5184	&2,979	&Train		\\ \cline{1-2}\cline{4-7}
I\tsb{33}& \cellcolor{festivals}				& \cellcolor{C16}	& 11.0 		&3456\x5184	&11,014	&Train	\\ \hline\hline
Total	& 4								& 16					&	 		&				&226,291&19/14	\\ \hline
\end{tabular}}
\squeezedown
\caption{DLR's Aerial Crowd Dataset consists of 33 aerial images from various scenes acquired through 16 flight campaigns, where different scene types and campaigns are denoted by different colors including \textbf{ALZ} Allianz Arena (Munich, Germany), \textbf{BOR} Signal Iduna Park (Dortmund, Germany), \textbf{MUC} Munich city center (Germany), \textbf{BRA} Braunschweig city center (Germany), \textbf{BAU} Bauma construction trade fair (Munich, Germany), \textbf{WIE} Oktoberfest (Munich, Germany), \textbf{RAR} Rock am Ring concert (Nuremberg, Germany), \textbf{STH} Southside music festival, \textbf{OAC} Open Air Concert (Germany), and \textbf{WIT} Church day (Wittenberg, Germany). The dataset is split into 19 training and 14 test images.}
\label{tab:dataset}
\end{table}
}

\newcommand{\crowdDatasetStatistics}{
\begin{table}
\centering
\resizebox{.95\columnwidth}{!}{\begin{tabular}{|c|c|c|c|c|c|c|c|} 
 \hline
   Dataset   			& Data				& Average     		& Number of	& Total Person 	& Average	& Maximum	& Minimum 	\\ 
   		   			& Collection			& Image Size (px)	& Images 	& Count	& Count		& Count		& Count 		\\
\hline\hline
   UCSD~\cite{Chan2008}				& CCTV cameras	& 158\x238			& 2,000		& 49,885	& 25		& 46		& 11		\\ \hline
   UCF\_CC\_50~\cite{Idrees2013}		& Web search		& 2101\x2888		& 50		& 63,974	& 1,279		& 4,633		& 94		\\ \hline
   WorldExpo'10~\cite{Zhang2015}		& CCTV cameras	& 576\x720			& 3,980		& 225,216	& 56		& 334		& 1			\\ \hline
   ShanghaiTech-A~\cite{Zhang2016} 	& Web search 		& 589\x868			& 482		& 241,677	& 501		& 3,139		& 33		\\ \hline
   ShanghaiTech-B~\cite{Zhang2016} 	& CCTV cameras	& 768\x1024			& 716		& 88,488	& 123		& 578		& 9			\\ \hline
   UCF-QNRF~\cite{Idrees2018}		& Web search		& 2013\x2902		& 1,535		& 1,251,642	& 815 		& 12,865	&	49		\\ \hline
   \rowcolor{rowColor}
   DLR-ACD				& Aerial imagery		& 3619\x5226		& 33		& 226,291	& 6,857		& 24,368	& 285						
\\ \hline	
 
\end{tabular}}
\squeezedown
\caption{Statistics of existing crowd datasets and the proposed DLR-ACD dataset.}
\label{tab:datasetStat}
\end{table}
}

\newcommand{\resultsShanghaiA}{
\begin{table}
\centering
\resizebox{.6\columnwidth}{!}{\begin{tabular}{|c|c|c|c|c|c|} 
 \hline
 Method                            &      Output         & \multicolumn{2}{c|}{Part-A}    &             \multicolumn{2}{c|}{Part-B} \\ \cline{3-6}                        
                          &  size   & MAE    & RMSE   & MAE & RMSE      \\ 
 \hline\hline
 MCNN~\cite{Zhang2016}          & 1:4   & 110.2 & 173.2 & 26.4  & 41.3       \\
 \hline
 CMTL~\cite{Sindagi2017}        & 1:1   & 101.3 & 152.4 & 20.0  & 31.1      \\
 \hline
 Switching CNN~\cite{Sam2017}   & 1:4   & 90.4 & 135.0 & 21.6  & 33.4     \\
 \hline
 MSCNN~\cite{Zeng2017}          & 1:4   & 83.8 & 127.4 & 17.7  & 30.2     \\
 \hline
 D-ConvNet-v1~\cite{Shi2018}    & 1:8   & 73.5 & 112.3 & 18.7  &  26.0    \\
 \hline
 IG-CNN~\cite{Sam2018}          & 1:4   & 72.5 & 118.2 & 13.6  &  21.1    \\
 \hline
 SCNet~\cite{Wang2018}          & 1:1   & 71.9  & 117.9  & 9.3  & 14.4     \\
 \hline
ic-CNN~\cite{Ranjan2018}        & 1:1   & 69.8  & 117.3  & 10.4  &  16.7    \\
 \hline
 CSRNet~\cite{Li2018}           & 1:8   & 68.2  & 115.0  & 10.6  &  16.0    \\
 \hline
 Liu~\etal~\cite{Liu2018b}      & 1:4   & 67.6  & 110.6  & 10.1  & 18.8     \\
 \hline
 SANet~\cite{Cao2018}           & 1:1   & 67.0  & 104.5  & \textbf{8.4}  & \textbf{13.6}   \\
 \hline
 \rowcolor{rowColor}
\XNET~(ours)	                        & 1:1   &  \textbf{66.2}     &   \textbf{102.0}   & 10.3  & 18.4    \\
 \hline
 
\end{tabular}}
\squeezedown
\caption{Crowd counting results on the ShanghaiTech dataset.}
\label{tab:shanghaiResults}
\end{table}
}

\newcommand{\resultsACD}{
\begin{table}
\centering
\resizebox{.7\columnwidth}{!}{\begin{tabular}{|l|c|c|c||c|c|c|c|} 
 \hline
 Method & MAE   & MNAE   & RMSE   & Precision & Recall  & F1-Score  \\ 
 \hline\hline
 MCNN       & 1989.7 & 0.87  & 3016.3   & 0.43  & 0.41  & 0.39  \\
 \hline
 ic-CNN     & 1481.3  & 0.72 & 2087  & 0.44  & \textbf{0.52}  & 0.46  \\
 \hline
 CSRNet     & 3388.8  & 0.71 & 4456.5  & 0.20  & 0.33  & 0.24   \\
 \hline
 Liu~\etal  & \textbf{833.3} & 0.25 & \textbf{1085.9}  & 0.45 & 0.44  & 0.44  \\
 \hline
\rowcolor{rowColor}
 \XNET~(ours) & 906.0 & \textbf{0.21}  & 1307.4 & \textbf{0.51}  & 0.48 & \textbf{0.49} \\
 \hline
 
\end{tabular}}
\squeezedown
\caption{Crowd counting and detection results on the DLR-ACD dataset.}
\label{tab:ACDResults}
\end{table}
}

\title{\XNET: Crowd Counting and Density Map Estimation in Aerial and Ground Imagery}

\addauthor{Reza Bahmanyar}{Reza.Bahmanyar@dlr.de}{1}
\addauthor{Eleonora Vig}{Eleonora.Vig@dlr.de}{1}
\addauthor{Peter Reinartz}{Peter.Reinartz@dlr.de}{1}

\addinstitution{
 Remote Sensing Technology Institute,\\
 German Aerospace Center (DLR),\\
 Wessling, Germany
}

\runninghead{Bahmanyar et al.}{\XNET \ -- Crowd Counting in Aerial Imagery}

\def\etal{\emph{et al}\bmvaOneDot}

\begin{document}

\maketitle

\begin{abstract}
In spite of the many advantages of aerial imagery for crowd monitoring and management at mass events, datasets of aerial images of crowds are still lacking in the field. As a remedy, in this work we introduce a novel crowd dataset, the DLR Aerial Crowd Dataset (DLR-ACD), which is composed of 33 large aerial images acquired from 16 flight campaigns over mass events with 226,291 persons annotated. To the best of our knowledge, DLR-ACD is the first aerial crowd dataset and will be released publicly. 
To tackle the problem of accurate crowd counting and density map estimation in aerial images of crowds, this work also proposes a new encoder-decoder convolutional neural network, the so-called Multi-Resolution Crowd Network (\XNET).
The encoder is based on the VGG-16 network and the decoder is composed of a set of bilinear upsampling and convolutional layers. Using two losses, one at an earlier level and another at the last level of the decoder, \XNET~estimates crowd counts and high-resolution crowd density maps as two different but interrelated tasks.
In addition, \XNET~utilizes contextual and detailed local information by combining high- and low-level features through a number of lateral connections inspired by the Feature Pyramid Network (FPN) technique.
We evaluated \XNET~on the proposed DLR-ACD dataset as well as on the ShanghaiTech dataset, a CCTV-based crowd counting benchmark. The results demonstrate that {\XNET}~outperforms the state-of-the-art crowd counting methods in estimating the crowd counts and density maps for both aerial and CCTV-based images.

\end{abstract}

%-------------------------------------------------------------------------
\section{Introduction}
\label{sec:intro}

\begin{figure}
    \centering
    \subfigure{\includegraphics[width=.425\textwidth]{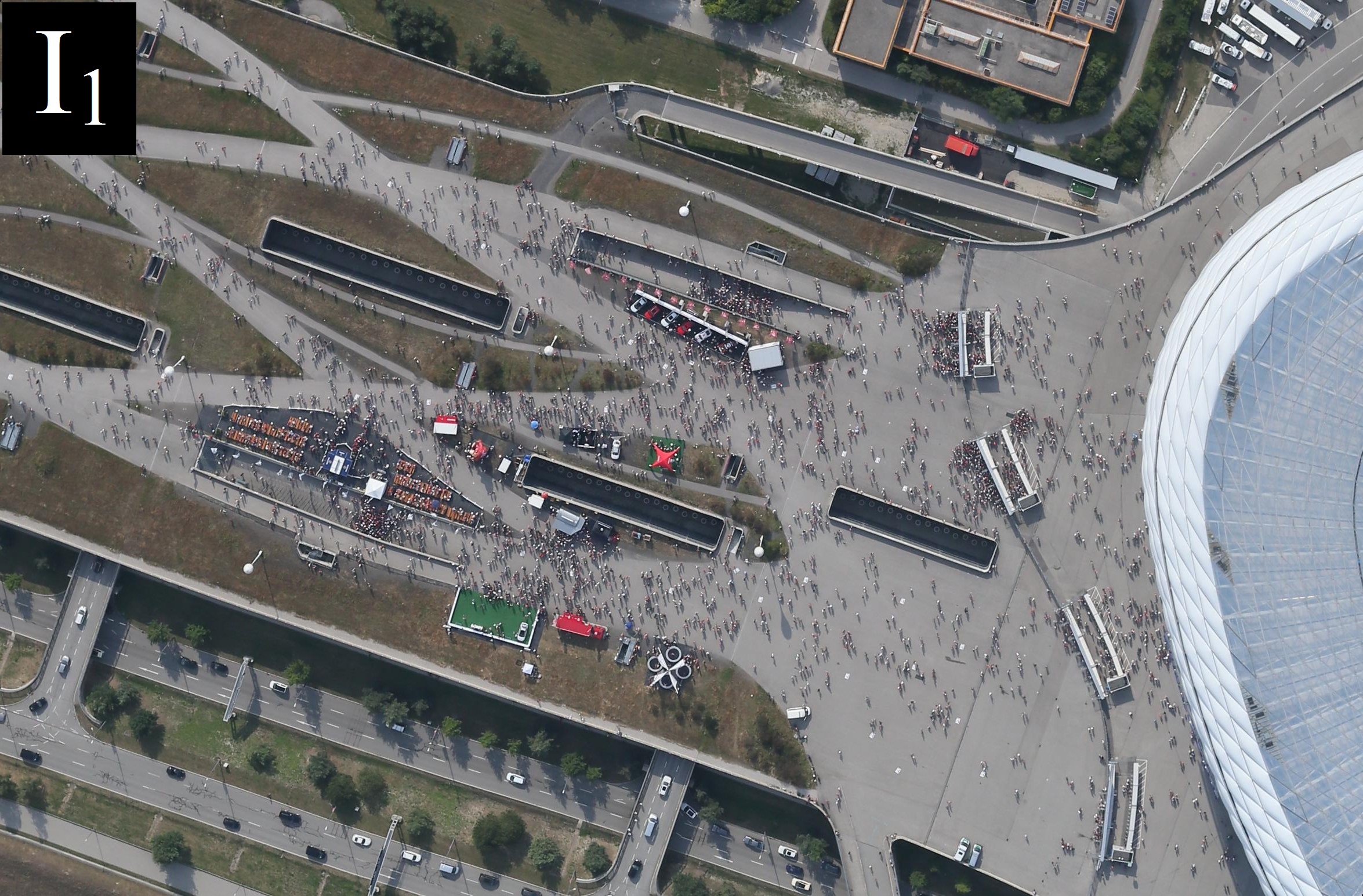}}
    \subfigure{\includegraphics[width=.42\textwidth]{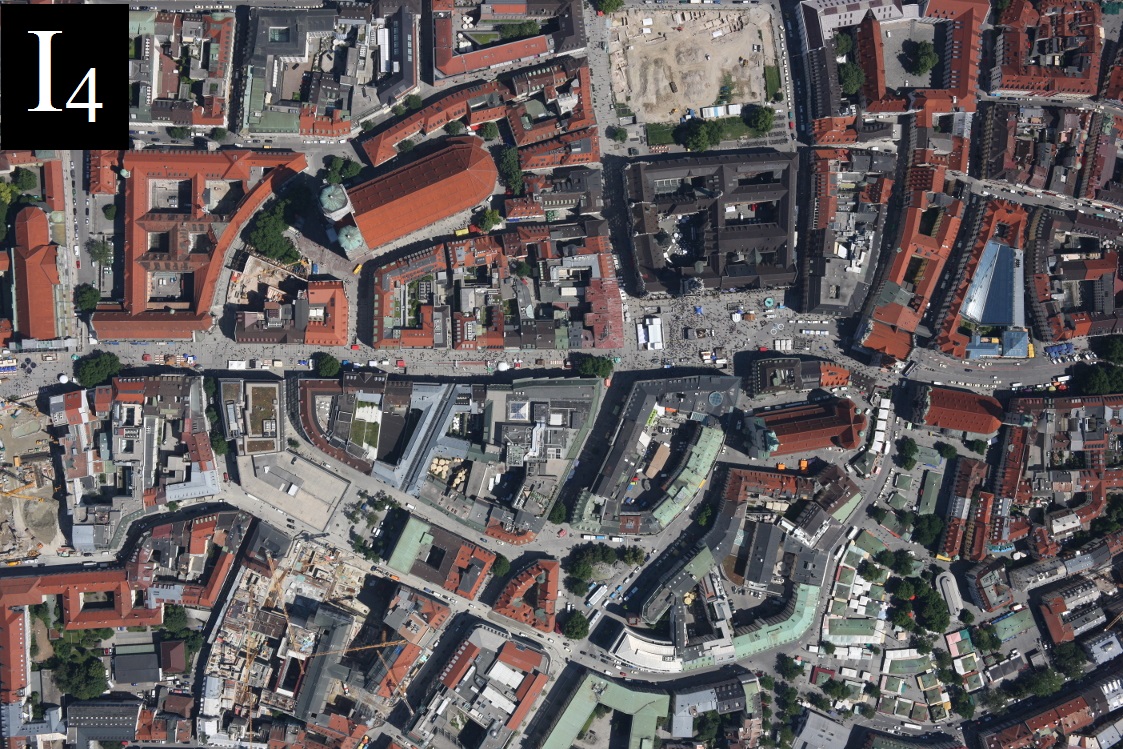}}
    \subfigure{\includegraphics[width=.421\textwidth]{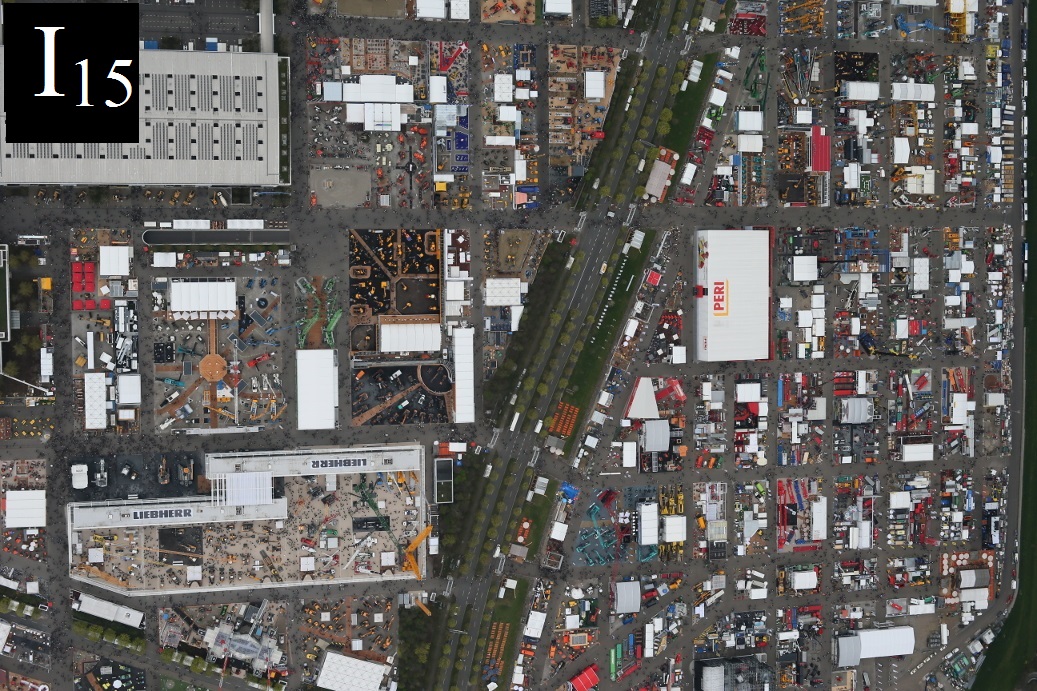}}
    \subfigure{\includegraphics[width=.421\textwidth]{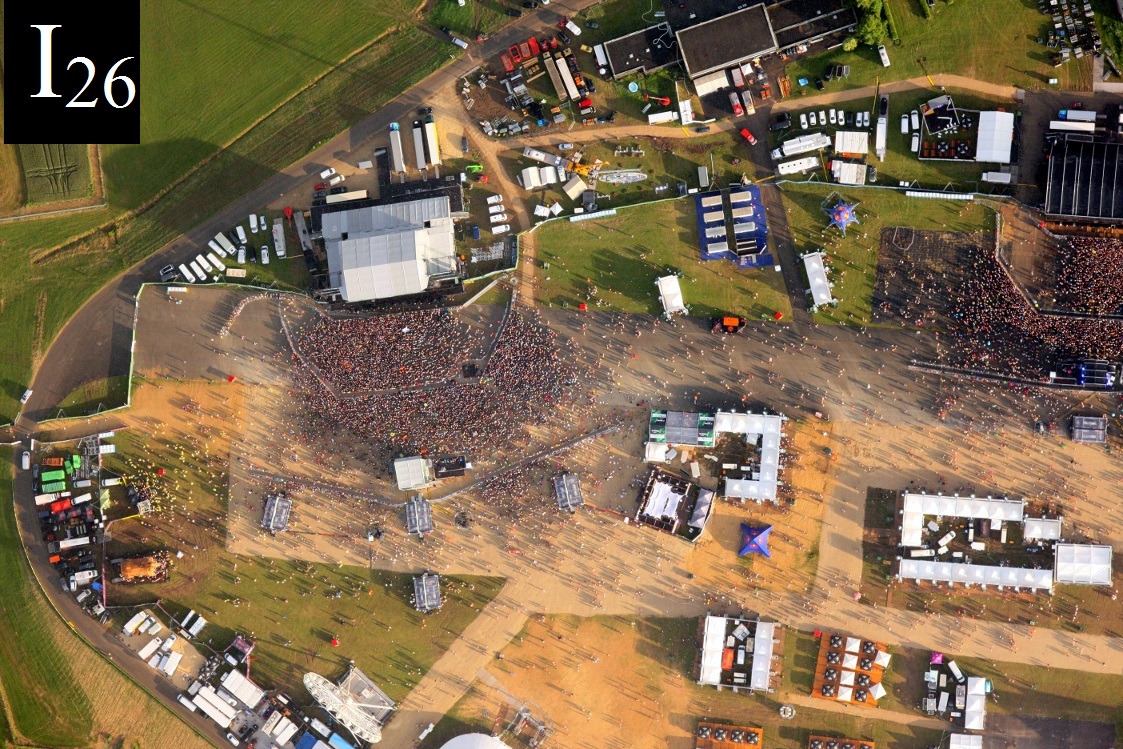}}
    \caption{Example images from the DLR-ACD dataset captured at a sport event, city center, a trade fair, and a concert. The image details can be found in Table~\ref{tab:dataset} using the given IDs.} 
    \label{fig:sampleData}
\end{figure}

Crowd counting and crowd density estimation play essential roles in safety monitoring and behavior analysis especially in the case of mass events. They can lead to early detection of congestion or security-related abnormalities informing and helping organizers and decision makers to avoid crowd disasters~\cite{Idrees2018}.
Closed-Circuit Television (CCTV) surveillance cameras have been conventionally used for crowd monitoring and they have become ubiquitous in recent years providing large number of images with various perspectives, scales, and illumination conditions.
However, for mass events spread over wide open areas with thousands of people attending, monitoring the crowd from above using aerial imagery (e.g., using airborne platforms) was shown to be advantageous due to the wider field of view and smaller occlusion effects as compared to CCTV images~\cite{Cui2017}.
Nevertheless, in spite of the increasing volume of available aerial images due to the advances in airborne and UAV platforms, crowd counting and density estimation datasets and methods for aerial imagery are still lacking in the domain.
Therefore, as one of the two main contributions of this work, we introduce a novel crowd dataset, the DLR Aerial Crowd Dataset (DLR-ACD), which is composed of 33 large aerial images (the average image size is 3619$\times$5226 pixels) acquired by standard DSLR cameras installed on an airborne platform on a helicopter. The images come from 16 flight campaigns, i.e.\ different mass events, and the dataset contains 226,291 person annotations. Figure~\ref{fig:sampleData} shows example images from DLR-ACD. To the best of our knowledge, DLR-ACD is the first aerial image crowd dataset and, with it, we hope to promote research on aerial crowd analysis. The dataset will be released at
{\small{\url{https://www.dlr.de/eoc/en/desktopdefault.aspx/tabid-12760/22294\_read-58354/}}}.

Despite the many benefits, crowd counting and density estimation is still a complex task in practice. For example, detecting and counting people in low resolution surveillance images, in which each person may only cover a few pixels and occlusion is frequent, is very difficult even for human experts~\cite{Kang2019,Liu2018}.
Therefore, developing automatic methods for precise crowd counting and density estimation is of high interest. 
In recent years, methods based on Convolutional Neural Networks (CNNs) have achieved promising results~\cite{Boominathan2016,Zhang2016,Sam2017,Sam2018,Sindagi2018,Zeng2017,Cao2018,Wang2018,Idrees2018,Huang2017,Liu2018b,Shi2018,Ranjan2018}. 
In contrast to using preset features, CNN-based methods learn features during the training which allows them to better cope with images with arbitrary perspectives, scales, and crowd densities~\cite{Sindagi2017}.

Considering the advantages of the existing crowd counting and density estimation methods and the new challenges presented by our aerial dataset such as extremely small objects and complex backgrounds, in this work, we propose Multi-Resolution Crowd Network (\XNET) which relies on a pre-trained VGG-16 model as an encoder and a combination of bilinear upsampling and convolution layers as a decoder. In addition, in order to
preserve as much high-resolution signal as possible while extracting multi-scale features,
the encoder and decoder are connected through lateral connections with element-wise addition similar to the Feature Pyramid Network (FPN) architecture~\cite{Lin2017}. 
While high-level features provide contextual information, the low-level features extract detailed information. In order to collect information from people with various sizes and in different crowd density conditions, \XNET~propagates information in all levels from the bottom to the top.
Taking advantage of this structure, \XNET~shows high robustness and transferability by achieving superior crowd counting and density map estimation results on both aerial and CCTV-based datasets. It outperforms the state-of-the-art methods on the ShanghaiTech dataset~\cite{Zhang2016}, a challenging CCTV-based benchmark, by achieving the smallest Mean Absolute count Error (MAE). In addition, on our proposed DLR-ACD dataset, \XNET~achieves the smallest counting MAE and the highest F1-score in person detection.

In summary, our main contributions are:
\squeezeup
\begin{itemize}
    \item We introduce a novel aerial crowd dataset, the so-called DLR-ACD.
    \squeezeup
    \item We propose \XNET~which is able to deal with the existing challenges in both aerial and CCTV-based crowd images.
    \squeezeup
    \item \XNET~achieves state-of-the-art results on the DLR-ACD and ShanghaiTech datasets.
\end{itemize}

\section{Related Works}\label{sec:relatedWorks} 
There is a vast literature on crowd counting and crowd density estimation in computer vision~\cite{Sindagi2018}. 
In order to deal with scale variation, a number of previous works proposed employing networks with multi-column architectures to obtain receptive fields with various sizes and extract features at different scales.
A three-column architecture developed by Zhang~\etal~\cite{Zhang2016}, the so-called MCNN, and a combination of shallow and deep networks proposed by Boominathan~\etal~\cite{Boominathan2016} are examples of multi-column CNNs.
As an approach toward improving the network performance in the presence of significant crowd density variations, Sam~\etal~\cite{Sam2017} proposed a switching multi-column CNN in which each column is trained separately on image patches of specific crowd densities. 
Ranjan~\etal~\cite{Ranjan2018} proposed ic-CNN, a two-branch network with a feed-forward structure, which incorporates low-resolution prediction maps for generating high-resolution crowd density maps.

Despite the improvement achieved by the multi-column approaches, their scale-invariance highly depends on the number of columns and their receptive field sizes~\cite{Sindagi2018}. Furthermore, their depths are usually limited as they present heavy computational overhead. Therefore, single-column architectures have been favoured in most of the recent crowd counting networks.
Zeng~\etal~\cite{Zeng2017} proposed a single column CNN, the so-called MSCNN, composed of scale aggregation blocks to tackle scale variations. Later Cao~\etal~\cite{Cao2018} used more sophisticated scale aggregation blocks in a deeper CNN together with a composition loss (Euclidean and local pattern consistency loss), the so-called SANet, and improved the count accuracy significantly.
Wang~\etal~\cite{Wang2018} also relied on scale aggregation blocks in designing scale aware residual modules for SCNet, a single column crowd counting CNN.
In order to tackle the variations in crowd density and appearance, Sam~\etal~\cite{Sam2018} proposed a growing CNN, the so-called IG-CNN, in which a base CNN is recursively split into two child-CNNs each becoming an expert on certain crowd types during training.

The advantages of multi-task CNNs have inspired Sindagi~\etal~\cite{Sindagi2017} to develop a cascaded multi-task CNN, the so-called CMTL, which simultaneously classifies crowds into different density levels and estimates density maps. 
Idrees~\etal~\cite{Idrees2018} developed a CNN that solves crowd counting, density estimation, and localization simultaneously. The network consists of a series of DenseNet~\cite{Huang2017} blocks with a composition of multiple intermediate losses, each optimizing the network for a ground-truth crowd density map smoothed with a different Gaussian kernel, and a final count loss.

A number of previous works took advantage of pre-trained networks as back-bones for crowd counting networks. In CSRNet, Li~\etal~\cite{Li2018} coupled VGG-16  with a dilated CNN as the back-end to obtain larger receptive fields and thus, better count accuracy. 
Later Liu~\etal~\cite{Liu2018b} combined the high and low level features by employing the VGG-16 network and FPN. 
This allows preserving and propagating the fine-grained details of small targets and also incorporating a large degree of context information.
Dealing with the over-fitting problem of crowd counting networks, Shi~\etal~\cite{Shi2018} proposed a learning strategy based on deep negative correlation learning applied to a modified VGG-16 network~\cite{Simonyan2015} which results in generalizable features through learning a pool of regressors to estimate the crowd density.

Our proposed \XNET~is based on a single-column encoder-decoder architecture similar to SegNet~\cite{Badrinarayanan2017}, U-Net~\cite{Ronneberger2015}, and Li~\etal~\cite{Liu2018b}. It uses a pre-trained VGG-16 as encoder. \XNET~is different from SegNet and U-Net in the decoder structure and the lateral connections. While the lateral connections of SegNet and U-Net are based on max-pooling indices and concatenations of earlier convolutional layers, respectively, \XNET~takes advantage of FPN-based lateral connections; however, it is different from~\cite{Liu2018b} in the number and wiring of the lateral connections and the decoder structure (e.g., the technique and the level of upsampling). It is different from~\cite{Li2018} in the number of the used convolutional blocks and the decoder structure. \XNET~considers crowd counting and density map estimation as two interrelated tasks and performs a multi-resolution prediction. It is different from the multi-task networks~\cite{Idrees2018,Sindagi2017} in the network structure and the task formulations.

\section{DLR's Aerial Crowd Dataset}

DLR's Aerial Crowd Dataset (DLR-ACD) is a collection of 33 large RGB aerial images with average size of 3619$\times$5226 pixels acquired through 16 different flight campaigns performed between 2011 and 2017. The aerial images were captured at various mass events and over urban scenes involving crowds, such as sport events, city centers, open-air fairs and festivals. The images were recorded using a camera system composed of three standard DSLR cameras (a nadir-looking and two side-looking cameras) mounted on an airborne platform installed on a helicopter flying at an altitude between 500\,m to 1600\,m. 
The different flight altitudes resulted in a range of spatial resolutions (or ground sampling distances -- GSD) from 4.5~cm/pixel to 15~cm/pixel and we also consider different viewing angles. Furthermore, the images were selected so that they represent different crowd densities and crowd behavior from the sparse moving crowds in city centers to the very dense (mostly) stationary ones at concerts. The dataset was labeled manually with point-annotations on individual people taking about 80 hours, and resulted in 226,291 person annotations, ranging from 285 to 24,368 annotations per image.
Crowd annotation in aerial images is a challenging task due to the large image sizes as well as the large number and the small size of the people in the images. While in dense crowd areas, discriminating each person from adjacent people is difficult, in sparse crowd areas localizing and discriminating each person from similar-looking objects is also challenging and time consuming.

\begin{figure}
    \centering
    \subfigure[]{\includegraphics[width=.64\textwidth]{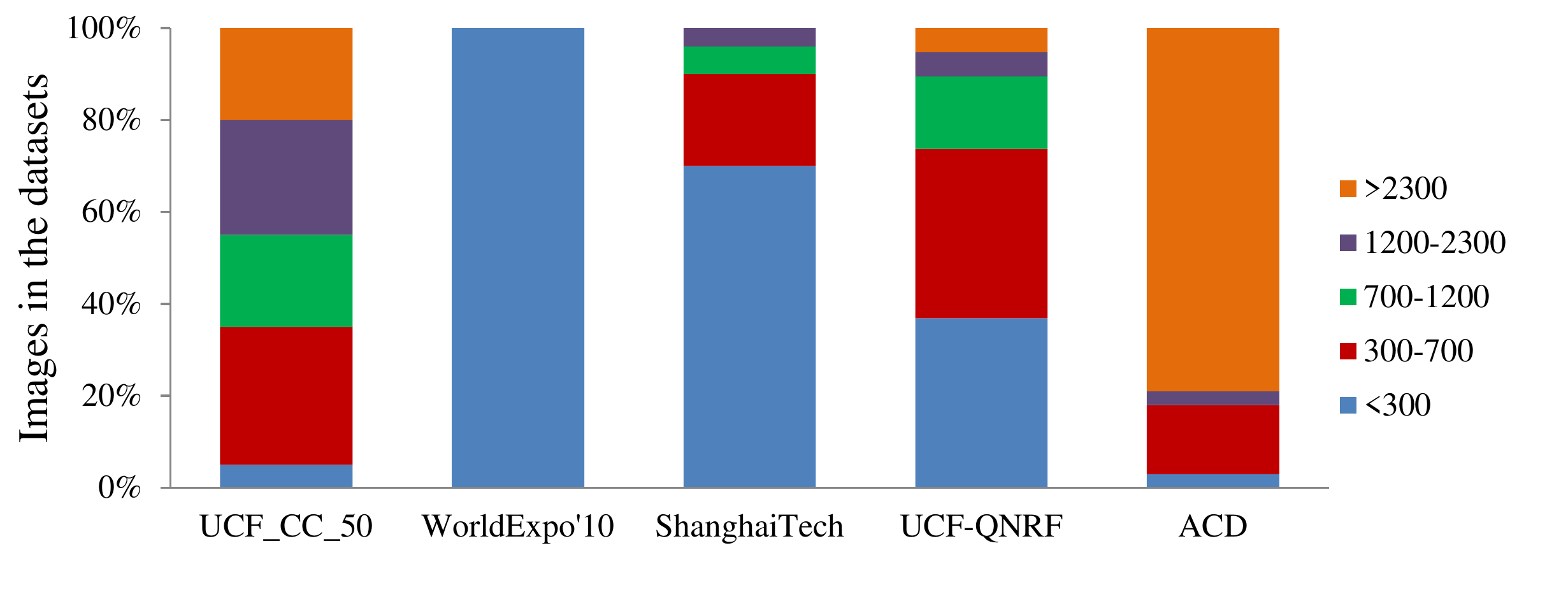}}
    \subfigure[]{\includegraphics[width=.34\textwidth]{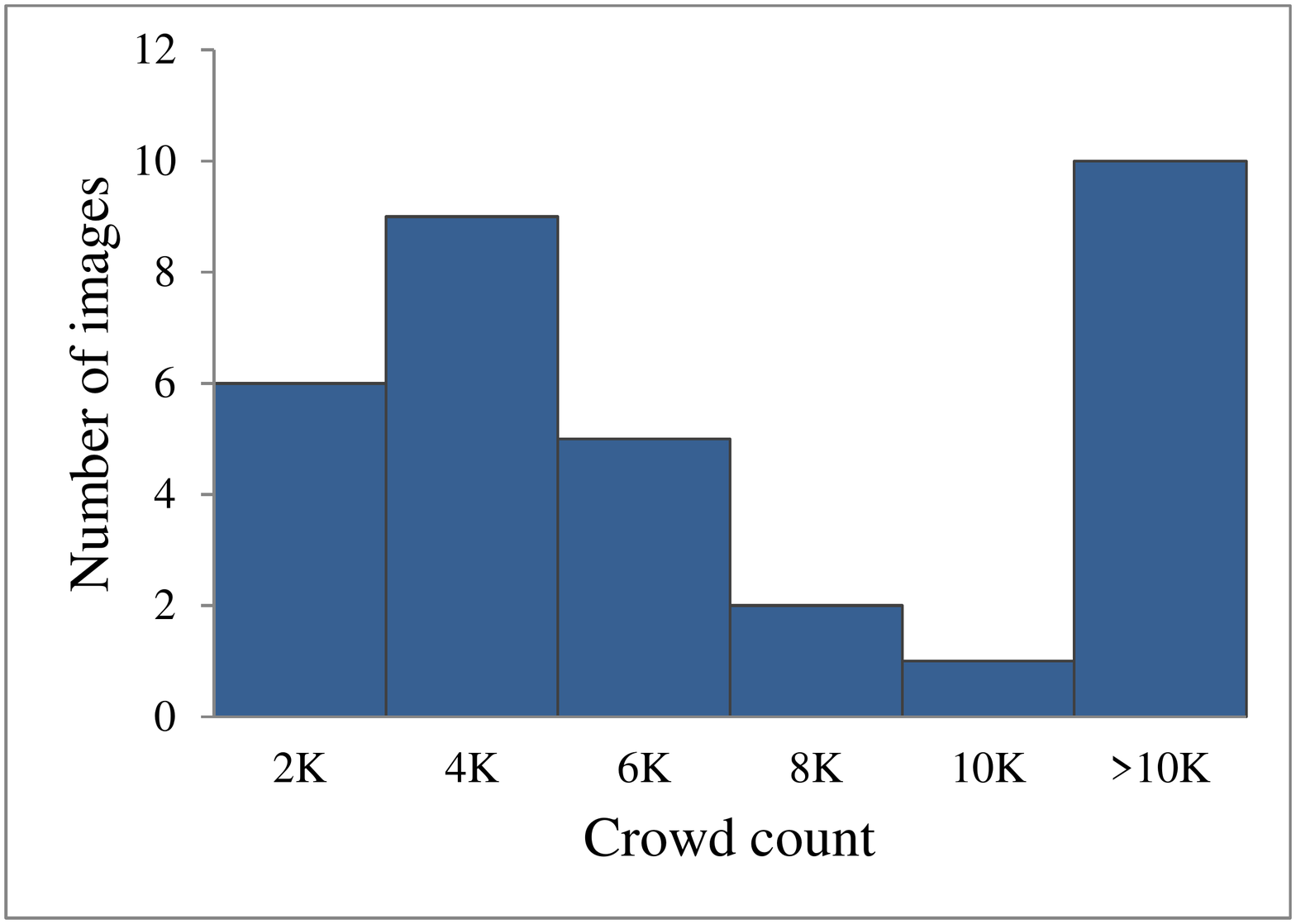}}
    \caption{(a) Distribution of per-image person counts of five datasets including our aerial crowd dataset (DLR-ACD). Graph inspired by~\cite{Idrees2018}. (b) Histogram of the DLR-ACD's crowd counts.}
    \label{fig:datasetStat}
\end{figure}

Table~\ref{tab:dataset} shows detailed information about the images and the annotations. Our images come from four types of events: sports events, fairs (e.g.\ trade fairs, Oktorberfest, etc.), and (music) festivals. To ensure that all scenes are covered in our train/test splits and that images from the same campaign are either in the training or in the test set, the dataset was manually split into 19 training and 14 test images, and the splits were not randomized.
The counts in the training and test sets are 138,151 and 88,140 persons.
%
%
%DLR-ACD details
\datasetDetailsTable
Table~\ref{tab:datasetStat} and Figure~\ref{fig:datasetStat} show the statistics of existing crowd datasets as well as our dataset. Among them, DLR-ACD is the first dataset that provides aerial views of crowds and therefore presents different challenges than existing datasets. Its images are much larger in size which might lead to higher computational costs and memory requirements. For example, when converted to match the image size of the widely-used ShanghaiTech-A dataset, DLR-ACD is $2.5\times$ larger.
In addition, as Figure~\ref{fig:datasetStat} shows, most of the images in DLR-ACD contain a large number of people ($>2K$) which is very different from the other crowd datasets. Furthermore, crowd densities vary significantly within and between images due to their large fields of view.
%

% Comparison table
\crowdDatasetStatistics

\section{\XNET}

\begin{figure}
    \centering
    \includegraphics[width=\textwidth]{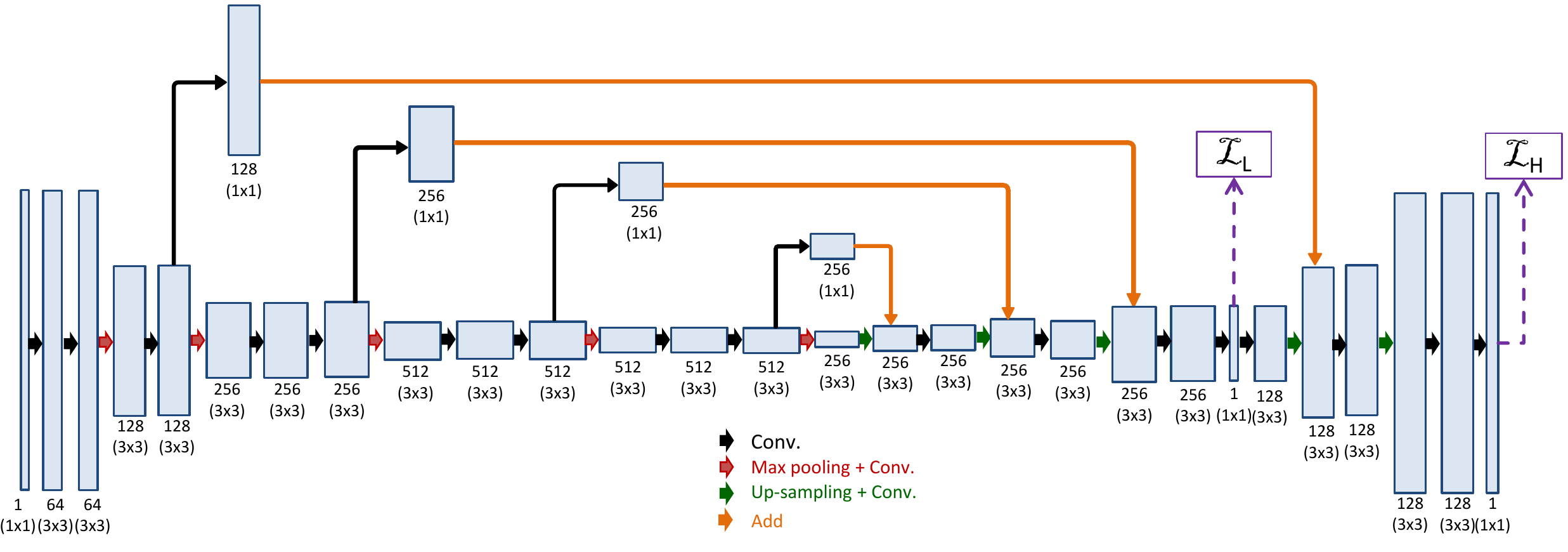}
    \caption{Structure of the proposed \XNET.}
    \label{fig:model}
\end{figure}

The Multi-Resolution Crowd Network (\XNET) utilizes an encoder-decoder structure to extract image features and generate crowd density maps. It takes a single image of arbitrary size and, in a fully-convolutional manner, predicts two density maps, one with 1/4 of the input image size for the people counting task and the other one with the input image size for the density map estimation task.  
For the encoder, \XNET~relies on a pre-trained VGG-16 network~\cite{Simonyan2015} (without batch normalization) composed of five CNN blocks, where the spatial size is reduced by half after each block using a max-pooling layer.
The decoder is composed of five CNN blocks, each preceded by an up-sampling layer based on bilinear interpolation which increases the spatial size by a factor of two. 
Figure~\ref{fig:model} illustrates the network structure. 
The number of feature maps and the used convolution kernel sizes are given below each box. After each convolution, ReLU nonlinearity is applied except for the layers with 1\x1 kernels.

Dealing with the diverse backgrounds of the crowd images (and of the aerial images in particular), using a deep CNN with multiple pooling layers helps reducing the influence of high frequency and small irrelevant background objects by increasing the receptive field size and extracting more contextual information. However, this could also remove the relevant details of the target objects (people) which is critical due to their small sizes. 
Therefore, \XNET~employs an FPN-based mechanism to combine the contextual information of the higher-level features and the detailed information provided by the lower-level features by element-wise adding the feature maps from the earlier stages to the ones in the later stages. This also helps avoiding the vanishing gradient problem. Using convolution layers with 1\x1 kernels on each lateral connection allows linear transformation and dimension reduction in the filter space. 

While most of the proposed methods focus on the counting task, \XNET~considers crowd counting and density map estimation as two interrelated tasks. To this end, being inspired by the effective composition of multiple losses in~\cite{Idrees2018,Ranjan2018,Sindagi2017}, \XNET~takes advantage of two losses at different resolutions for the counting and density map estimation tasks.
It has been shown by a number of previous works~\cite{Zhang2016,Zeng2017,Sam2018,Liu2018b} that crowd counting could be performed without upsampling the prediction maps to the size of input images. This reduces the prediction complexity as the fine details do not have to be predicted. Taking this into account, \XNET~generates a low-resolution prediction (1/4 times smaller than the input image) at an earlier stage of the decoder and compares it to a downsampled ground-truth, optimizing the $\mathcal{L}_L$ pixel-wise Mean Squared Error (MSE) loss. The network is supposed to predict the image count in this stage and should use the rest of the decoder for predicting the full-resolution crowd density maps, with a high localization precision, while keeping the count close to the ground truth by optimizing the $\mathcal{L}_H$ loss, another pixel-wise MSE loss. MSE is widely used by crowd counting networks. The total loss is then computed as: $\mathcal{L}_{total} = \mathcal{L}_L + \lambda\mathcal{L}_H $, where $\lambda$ is empirically set to 0.0001. The number of network parameters is 20.3~M.

\section{Results and Discussion}
In this section, \XNET~is evaluated on the DLR-ACD and ShanghaiTech datasets. In the training, the Adam optimization algorithm with a learning rate of $3e^{-6}$ was employed and the batch sizes were empirically set to 60 and 40 for the DLR-ACD and ShanghaiTech datasets, respectively. In addition, apart from the VGG's parameters, all network parameters were randomly initialized by a Gaussian distribution with a zero mean and a standard deviation of 0.01.

\subsection{Evaluation Metrics}

Mean Absolute Error (MAE) and Root Mean Squared Error (RMSE) are the metrics that have been widely used for evaluating crowd counting performance.
However, these metrics treat all images equally without considering the differences between their true counts. This could be problematic for the datasets with large variations of the image counts. For example, in the DLR-ACD dataset, I$_{10}$ and I$_{19}$ contain 285 and 24,368 people, respectively. If the prediction for each image has a count error of 1K, MAE and RMSE consider them as equal errors for both images. Nevertheless, 1K for I$_{10}$ is about 350\% off its true count whereas for I$_{19}$ it is only about 4\%. Thus, in order to represent model performances on each image in single-image crowd counting scenarios, evaluation metrics should also consider the image differences such as in their counts. In this work, for analyzing model counting performance on the DLR-ACD dataset, in addition to the conventional metrics, we use Mean Normalized Absolute Error (MNAE), that evaluates the predictions for each image differently by considering the image's true count. Assuming $C_i$ and $\hat{C}_i$ as the ground-truth and predicted counts for image $i$ ($i\in\{1,2, ...,N\}$), respectively, the metrics are computed as:
\begin{equation}
    MAE=\frac{1}{N}\sum_{i=1}^{N}|C_i - \hat{C}_i|,\;\;\;\; MNAE = \frac{1}{N}\sum_{i=1}^{N}\frac{|C_i - \hat{C}_i|}{C_i},\;\;\;\; RMSE=\sqrt{\frac{1}{N}\sum_{i=1}^{N}(C_i - \hat{C}_i)^2}.
\end{equation}

For person detection, people locations are extracted by detecting local maxima in the predicted high-resolution density maps. For \XNET, the number of people (or maxima) to be extracted is given by the person count estimated from the low-resolution prediction (i.e.\ the output of the earlier stage of the decoder). 
To evaluate the detections, we use the standard precision, recall, and F1-score. A detection is counted as true positive, if it lies within half a meter of a ground-truth person location, which is quite a strict requirement. Here, we assume that the GSD (in pixel/m) of the aerial image is known at test time. 

\subsection{Experiments on the DLR-ACD dataset}
In order to generate the ground truth density maps, we adopt the standard approach of smoothing the person locations with a small 2D Gaussian~\cite{Zhang2016}, a procedure which is aimed at avoiding the imbalance between the number of positive and negative samples (a pixel on each person versus all other image pixels). 
As the spatial resolution (GSD) of our aerial images are known and the distortions caused by the homography between the ground and image planes are negligible, the Gaussian smoothing is adapted according to the GSDs of the images. To this end, assuming that the area covered by each person (looking from above) is roughly a square of 0.5\x0.5~m, the standard deviation (in pixels) of the Gaussian is computed as: $\sigma=\lfloor\frac{1}{3}\frac{0.5}{GSD}\rfloor$, where the GSD is in meters and each person will lie within 3$\sigma$ of the Gaussian distribution. Since the Gaussian kernels being used are normalized ($\int G_\sigma(x)=1$), a sum over the ground-truth density map gives the total person count in the image.

For training, the images were tiled evenly into patches of 320\x320 pixels with 50\% overlap, which resulted in 11,908 patches. Then, from each patch, two samples of size 256\x256 were randomly cropped, where one sample was used as it was and the other was randomly augmented. For the augmentation, three rotations (90$^{\circ}$, 180$^{\circ}$, and 270$^{\circ}$), two flips (left-right and up-down), and two scaling (up- and downsample) were considered on a random basis.

\begin{figure}
    \centering
    \subfigure{\includegraphics[width=.23\textwidth]{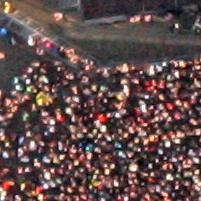}}
    \subfigure{\includegraphics[width=.23\textwidth]{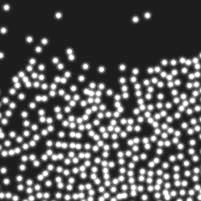}}
    \subfigure{\includegraphics[width=.23\textwidth]{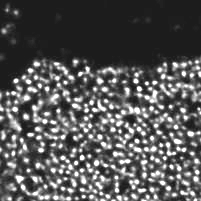}}    \subfigure{\includegraphics[width=.23\textwidth]{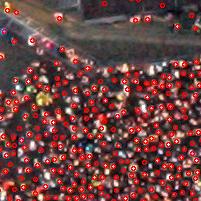}}
    
    %%%%%%%
    \subfigure{\includegraphics[width=.23\textwidth]{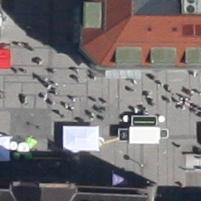}}
    \subfigure{\includegraphics[width=.23\textwidth]{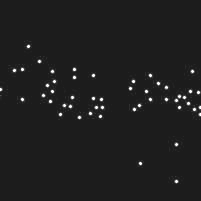}}
    \subfigure{\includegraphics[width=.23\textwidth]{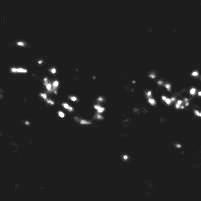}}
    \subfigure{\includegraphics[width=.23\textwidth]{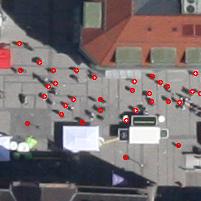}}
    \caption{\XNET's results on the DLR-ACD dataset. Each row (from left to right) shows a sample cropped image, its ground truth, the estimated density map, and detection results.}
    \label{fig:QResultsACD}
\end{figure}

As the qualitative results in Figure~\ref{fig:QResultsACD} show, \XNET~performs well in estimating high-resolution crowd maps in both dense and sparse crowd scenarios. However, it misses some people due to background clutter. Furthermore, the quantitative results of Table~\ref{tab:ACDResults} demonstrate that \XNET~outperforms other methods by a better count estimation (lowest MNAE) and a higher quality of the estimated density maps for detection tasks (highest F1-score).
In addition, considering the \XNET's number of parameters (20.3~M), its average inference time is 0.03~ms per image patch of 256\x256 pixels.  
\resultsACD

\subsection{Experiments on the ShanghaiTech dataset}
We also trained and validated \XNET~on the ShanghaiTech crowd dataset~\cite{Zhang2016}, which is one of the most widely used crowd benchmarks. This dataset is composed of two parts: Part-A contains 482 images (300 training and 182 test images) and Part-B contains 716 images (400 training and 316 test images). Statistics about this dataset can be seen in Table~\ref{tab:datasetStat} and Figure~\ref{fig:datasetStat}. In order to generate ground truth density maps, we followed the approach proposed in~\cite{Zhang2016}. In order to avoid over-fitting, we randomly cropped 20 patches of size 224\x224 from each training image. Then, as data augmentation, we applied left-right flipping to 30\% of the patches on a random basis. In addition, the images were converted into gray scale.

Table~\ref{tab:shanghaiResults} shows crowd counting performance of \XNET~compared to the state of the art on the ShanghaiTech dataset. \XNET~outperforms all other methods on Part-A by achieving the lowest MAE and RMSE values, and achieves competitive results on Part-B.  
\resultsShanghaiA

\section{Conclusion}
This work proposed Multi-Resolution Crowd Network (\XNET), a convolutional neural network for accurate crowd counting and density map estimation in aerial and ground imagery. \XNET~considers crowd counting and density map estimation as two interrelated tasks addressed at different resolutions. In addition, a novel aerial crowd dataset, the so-called DLR-ACD, was introduced which  promotes crowd monitoring and management from aerial imagery. The superior performance of \XNET~on the DLR-ACD and ShanghaiTech (a ground imagery benchmark) datasets were shown through quantitative and qualitative results. Furthermore, results demonstrated that the estimated crowd maps can be used also for person detection thanks to their high-resolution and accuracy. 

%

%\newpage
\bibliography{egbib}

\begin{thebibliography}{25}
\providecommand{\natexlab}[1]{#1}
\providecommand{\url}[1]{\texttt{#1}}
\expandafter\ifx\csname urlstyle\endcsname\relax
  \providecommand{\doi}[1]{doi: #1}\else
  \providecommand{\doi}{doi: \begingroup \urlstyle{rm}\Url}\fi

\bibitem[{Badrinarayanan} et~al.(2017){Badrinarayanan}, {Kendall}, and
  {Cipolla}]{Badrinarayanan2017}
V.~{Badrinarayanan}, A.~{Kendall}, and R.~{Cipolla}.
\newblock {SegNet}: A deep convolutional encoder-decoder architecture for image
  segmentation.
\newblock \emph{IEEE Transactions on Pattern Analysis and Machine
  Intelligence}, 39\penalty0 (12):\penalty0 2481--2495, Dec 2017.

\bibitem[Boominathan et~al.(2016)Boominathan, Kruthiventi, and
  Babu]{Boominathan2016}
L.~Boominathan, S.~Kruthiventi, and R.~V. Babu.
\newblock {CrowdNet: A Deep Convolutional Network for Dense Crowd Counting}.
\newblock In \emph{Proceedings of the ACM International Conference on
  Multimedia (MM)}, pages 640--644, 2016.

\bibitem[Cao et~al.(2018)Cao, Wang, Zhao, and Su]{Cao2018}
X.~Cao, Z.~Wang, Y.~Zhao, and F.~Su.
\newblock Scale aggregation network for accurate and efficient crowd counting.
\newblock In \emph{Proceedings of the European Conference on Computer Vision
  (ECCV)}, 2018.

\bibitem[{Chan} et~al.(2008){Chan}, Liang, and {Vasconcelos}]{Chan2008}
A.~B. {Chan}, J.~Liang, and N.~{Vasconcelos}.
\newblock Privacy preserving crowd monitoring: Counting people without people
  models or tracking.
\newblock In \emph{Proceedings of IEEE Conference on Computer Vision and
  Pattern Recognition (CVPR)}, pages 1--7, 2008.

\bibitem[Cui et~al.(2017)Cui, {Meynberg}, and {Reinartz}]{Cui2017}
S.~Cui, O.~{Meynberg}, and P.~{Reinartz}.
\newblock Bayesian linear regression for crowd density estimation in aerial
  images.
\newblock In \emph{Proceedings of Joint Urban Remote Sensing Event (JURSE)},
  pages 1--4, 2017.

\bibitem[{Huang} et~al.(2017){Huang}, {Liu}, v.~d. {Maaten}, and
  {Weinberger}]{Huang2017}
G.~{Huang}, Z.~{Liu}, L.~v.~d. {Maaten}, and K.~Q. {Weinberger}.
\newblock Densely connected convolutional networks.
\newblock In \emph{Proceedings of IEEE Conference on Computer Vision and
  Pattern Recognition (CVPR)}, pages 2261--2269, 2017.

\bibitem[{Idrees} et~al.(2013){Idrees}, {Saleemi}, {Seibert}, and
  {Shah}]{Idrees2013}
H.~{Idrees}, I.~{Saleemi}, C.~{Seibert}, and M.~{Shah}.
\newblock Multi-source multi-scale counting in extremely dense crowd images.
\newblock In \emph{Proceedings of IEEE Conference on Computer Vision and
  Pattern Recognition (CVPR)}, pages 2547--2554, 2013.

\bibitem[Idrees et~al.(2018)Idrees, Tayyab, Athrey, Zhang, Al{-}M{\'{a}}adeed,
  Rajpoot, and Shah]{Idrees2018}
H.~Idrees, M.~Tayyab, K.~Athrey, D.~Zhang, S.~Al{-}M{\'{a}}adeed, N.~M.
  Rajpoot, and M.~Shah.
\newblock Composition loss for counting, density map estimation and
  localization in dense crowds.
\newblock In \emph{Proceedings of the European Conference on Computer Vision
  (ECCV)}, 2018.

\bibitem[{Kang} et~al.(2019){Kang}, {Ma}, and {Chan}]{Kang2019}
D.~{Kang}, Z.~{Ma}, and A.~B. {Chan}.
\newblock {Beyond Counting: Comparisons of Density Maps for Crowd Analysis
  Tasks -- Counting, Detection, and Tracking}.
\newblock \emph{IEEE Transactions on Circuits and Systems for Video
  Technology}, 29\penalty0 (5):\penalty0 1408--1422, May 2019.

\bibitem[{Li} et~al.(2018){Li}, {Zhang}, and {Chen}]{Li2018}
Y.~{Li}, X.~{Zhang}, and D.~{Chen}.
\newblock {CSRNet}: Dilated convolutional neural networks for understanding the
  highly congested scenes.
\newblock In \emph{Proceedings of IEEE/CVF Conference on Computer Vision and
  Pattern Recognition (CVPR)}, pages 1091--1100, 2018.

\bibitem[{Lin} et~al.(2017){Lin}, {Doll\'ar}, {Girshick}, {He}, {Hariharan},
  and {Belongie}]{Lin2017}
T.~{Lin}, P.~{Doll\'ar}, R.~{Girshick}, K.~{He}, B.~{Hariharan}, and
  S.~{Belongie}.
\newblock {Feature Pyramid Networks for Object Detection}.
\newblock In \emph{Proceedings of IEEE Conference on Computer Vision and
  Pattern Recognition (CVPR)}, pages 936--944, 2017.

\bibitem[Liu et~al.(2018)Liu, Jiang, Guo, Wang, and Liu]{Liu2018b}
M.~Liu, J.~Jiang, Z.~Guo, Z.~Wang, and Y.~Liu.
\newblock Crowd counting with fully convolutional neural network.
\newblock In \emph{Proceedings of IEEE International Conference on Image
  Processing (ICIP)}, pages 953--957, 2018.

\bibitem[{Liu} et~al.(2018){Liu}, {van de Weijer}, and {Bagdanov}]{Liu2018}
X.~{Liu}, J.~{van de Weijer}, and A.~D. {Bagdanov}.
\newblock Leveraging unlabeled data for crowd counting by learning to rank.
\newblock In \emph{Proceedings of IEEE/CVF Conference on Computer Vision and
  Pattern Recognition (CVPR)}, pages 7661--7669, 2018.

\bibitem[Ranjan et~al.(2018)Ranjan, Le, and Hoai]{Ranjan2018}
V.~Ranjan, H.~Le, and M.~Hoai.
\newblock Iterative crowd counting.
\newblock In \emph{Proceedings of the European Conference on Computer Vision
  (ECCV)}, pages 278--293, 2018.

\bibitem[Ronneberger et~al.(2015)Ronneberger, Fischer, and
  Brox]{Ronneberger2015}
O.~Ronneberger, P.~Fischer, and T.~Brox.
\newblock U-net: Convolutional networks for biomedical image segmentation.
\newblock In \emph{Proceedings of Medical Image Computing and Computer-Assisted
  Intervention (MICCAI)}, pages 234--241, 2015.

\bibitem[{Sam} et~al.(2017){Sam}, {Surya}, and {Babu}]{Sam2017}
D.~B. {Sam}, S.~{Surya}, and R.~V. {Babu}.
\newblock Switching convolutional neural network for crowd counting.
\newblock In \emph{Proceedings of IEEE Conference on Computer Vision and
  Pattern Recognition (CVPR)}, pages 4031--4039, 2017.

\bibitem[{Sam} et~al.(2018){Sam}, {Sajjan}, {Babu}, and {Srinivasan}]{Sam2018}
D.~B. {Sam}, N.~N. {Sajjan}, R.~V. {Babu}, and M.~{Srinivasan}.
\newblock Divide and grow: Capturing huge diversity in crowd images with
  incrementally growing {CNN}.
\newblock In \emph{Proceedings of IEEE/CVF Conference on Computer Vision and
  Pattern Recognition (CVPR)}, pages 3618--3626, 2018.

\bibitem[{Shi} et~al.(2018){Shi}, {Zhang}, {Liu}, {Cao}, {Ye}, {Cheng}, and
  {Zheng}]{Shi2018}
Z.~{Shi}, L.~{Zhang}, Y.~{Liu}, X.~{Cao}, Y.~{Ye}, M.~{Cheng}, and G.~{Zheng}.
\newblock Crowd counting with deep negative correlation learning.
\newblock In \emph{Proceedings of IEEE/CVF Conference on Computer Vision and
  Pattern Recognition (CVPR)}, pages 5382--5390, 2018.

\bibitem[Simonyan and Zisserman(2015)]{Simonyan2015}
K.~Simonyan and A.~Zisserman.
\newblock Very deep convolutional networks for large-scale image recognition.
\newblock In \emph{Proceedings of International Conference on Learning
  Representations (ICLR)}, 2015.

\bibitem[{Sindagi} and {Patel}(2017)]{Sindagi2017}
V.~A. {Sindagi} and V.~M. {Patel}.
\newblock {CNN}-based cascaded multi-task learning of high-level prior and
  density estimation for crowd counting.
\newblock In \emph{Proceedings of IEEE International Conference on Advanced
  Video and Signal Based Surveillance (AVSS)}, pages 1--6, 2017.

\bibitem[Sindagi and Patel(2018)]{Sindagi2018}
V.~A. Sindagi and V.~M. Patel.
\newblock A survey of recent advances in {CNN-based} single image crowd
  counting and density estimation.
\newblock \emph{Pattern Recognition Letters}, 107:\penalty0 3 -- 16, 2018.

\bibitem[Wang et~al.(2018)Wang, Xiao, Xie, Qiu, Zhen, and Cao]{Wang2018}
Z.~Wang, Z.~Xiao, K.~Xie, Q.~Qiu, X.~Zhen, and X.~Cao.
\newblock In defense of single-column networks for crowd counting.
\newblock In \emph{Proceedings of British Machine Vision Conference (BMVC)},
  2018.

\bibitem[{Zeng} et~al.(2017){Zeng}, {Xu}, {Cai}, {Qiu}, and {Zhang}]{Zeng2017}
L.~{Zeng}, X.~{Xu}, B.~{Cai}, S.~{Qiu}, and T.~{Zhang}.
\newblock Multi-scale convolutional neural networks for crowd counting.
\newblock In \emph{Proceedings of IEEE International Conference on Image
  Processing (ICIP)}, pages 465--469, 2017.

\bibitem[Zhang et~al.(2015)Zhang, Li, Wang, and Yang]{Zhang2015}
C.~Zhang, H.~Li, X.~Wang, and X.~Yang.
\newblock Cross-scene crowd counting via deep convolutional neural networks.
\newblock In \emph{Proceedings of IEEE Conference on Computer Vision and
  Pattern Recognition (CVPR)}, pages 833--841, 2015.

\bibitem[{Zhang} et~al.(2016){Zhang}, {Zhou}, {Chen}, {Gao}, and
  {Ma}]{Zhang2016}
Y.~{Zhang}, D.~{Zhou}, S.~{Chen}, S.~{Gao}, and Y.~{Ma}.
\newblock Single-image crowd counting via multi-column convolutional neural
  network.
\newblock In \emph{Proceedings of IEEE Conference on Computer Vision and
  Pattern Recognition (CVPR)}, pages 589--597, 2016.

\end{thebibliography}
\end{document}